\documentclass[letterpaper, 10 pt, journal, twoside]{IEEEtran}

\usepackage{graphicx}
\usepackage{hyperref}
\usepackage{balance}
\usepackage{tabularx}
\usepackage{multirow}
\usepackage{multicol}
\usepackage{booktabs}
\usepackage{caption}
\usepackage{mathtools}
\usepackage{siunitx}
\usepackage{amsmath}
\usepackage{MnSymbol}

\usepackage{amssymb}
\usepackage{kotex}
\usepackage{graphicx}
\usepackage{subcaption}
\usepackage{tablefootnote}
\usepackage{bbding}
\usepackage{threeparttable} 
\hypersetup{
    colorlinks=False, 
    urlcolor=black,
}
\usepackage{amsfonts}
\usepackage{algorithmic}
\usepackage{array}
\usepackage{textcomp}
\usepackage{stfloats}
\usepackage{url}
\usepackage{verbatim}
\usepackage{graphicx}
\def\BibTeX{{\rm B\kern-.05em{\sc i\kern-.025em b}\kern-.08em
    T\kern-.1667em\lower.7ex\hbox{E}\kern-.125emX}}

\begin{document}
\markboth{IEEE Robotics and Automation Letters. Preprint Version. Accepted APRIL, 2024}
{KIM \MakeLowercase{\textit{et al.}}: GRIL-Calib: Targetless Ground Robot IMU-LiDAR Extrinsic Calibration Method} 

\title{GRIL-Calib: Targetless Ground Robot IMU-LiDAR Extrinsic Calibration Method using Ground Plane Motion Constraints}

\author{TaeYoung Kim$^{1}$, Gyuhyeon Pak$^{2}$ and Euntai Kim$^{2, *}$
\thanks{Manuscript received: December, 17, 2023; Revised February, 28, 2024; Accepted April, 23, 2024.}
\thanks{This paper was recommended for publication by Editor Pascal Vasseur upon evaluation of the Associate Editor and Reviewers' comments. This work was partly supported by Institute of Information \& communications Technology Planning \& Evaluation (IITP) grant funded by the Korea government(MSIT) (No.2022-0-01025, Development of core technology for mobile manipulator for 5G edge-based transportation and manipulation) and Korea Evaluation Institute of Industrial Technology(KEIT) grant funded by the Korea government(MOTIE) (No.20023455, Development of Cooperate Mapping, Environment Recognition and Autonomous Driving Technology for Multi Mobile Robots Operating in Large-scale Indoor Workspace). \textit{(Corresponding author: Euntai Kim.)}}
\thanks{$^{1}$TaeYoung Kim is with the Department of Vehicle Convergence Engineering, Yonsei University, Seoul 03722, South Korea {\tt\footnotesize  tyoung96@yonsei.ac.kr}}%
\thanks{$^{2}$Gyuhyeon Pak and Euntai Kim are with the Department of Electrical and Electronic
Engineering, Yonsei University, Seoul 03722, South Korea {\tt\footnotesize \{ghpak, etkim\}@yonsei.ac.kr}}%
\thanks{Digital Object Identifier (DOI): see top of this page.}
}

\maketitle

\begin{abstract}

Targetless IMU-LiDAR extrinsic calibration methods are gaining significant attention as the importance of the IMU-LiDAR fusion system increases. Notably, existing calibration methods derive calibration parameters under the assumption that the methods require full motion in all axes. When IMU and LiDAR are mounted on a ground robot the motion of which is restricted to planar motion, existing calibration methods are likely to exhibit degraded performance. To address this issue, we present \textit{GRIL-Calib}: a novel targetless Ground Robot IMU-LiDAR Calibration method. Our proposed method leverages ground information to compensate for the lack of unrestricted full motion. First, we propose LiDAR Odometry (LO) using ground plane residuals to enhance calibration accuracy. Second, we propose the Ground Plane Motion (GPM) constraint and incorporate it into the optimization for calibration, enabling the determination of full 6-DoF extrinsic parameters, including theoretically unobservable direction. Finally, unlike baseline methods, we formulate the calibration not as sequential two optimizations but as a single optimization (SO) problem, solving all calibration parameters simultaneously and improving accuracy.
We validate our \textit{GRIL-Calib} by applying it to various real-world datasets and comparing its performance with that of existing state-of-the-art methods in terms of accuracy and robustness. Our code is available at \href{https://github.com/Taeyoung96/GRIL-Calib}{https://github.com/Taeyoung96/GRIL-Calib}.
\end{abstract}


\section{INTRODUCTION}

\IEEEPARstart{L}{iDAR} (Light Detection And Ranging) acquires more accurate 3D information than other sensors, making it broadly applicable to a variety of robotic applications such as ego-motion estimation \cite{vizzo2023kiss}, localization \cite{wang20233d} and mapping \cite{zhang2014loam} system. However, LiDAR has the drawback that when it is applied to fast motion, it is vulnerable to motion distortion.
Recently, to overcome the drawback, LiDAR has been used
in conjunction with IMU (Inertial Measurement Unit) \cite{xu2022fast}, \cite{shan2020lio}, \cite{bai2022faster}. The IMU effectively compensates for the motion distortion of the LiDAR, as it measures acceleration and angular
velocity at a high frequency. Obviously, accurate extrinsic calibration is a fundamental requirement in the IMU-LiDAR fusion system. Extrinsic calibration refers to the process of establishing the spatial relationship, including both translation and rotation, between multiple sensors. This process ensures that measurements from different sensors are accurately interpreted using the same coordinate frames. If the extrinsic calibration is not accurate enough, the fusion of IMU and LiDAR becomes less effective, leading to significant degradation in overall performance \cite{10106780}. 

\begin{figure}[t]
\centering
\includegraphics[width=0.85\columnwidth]{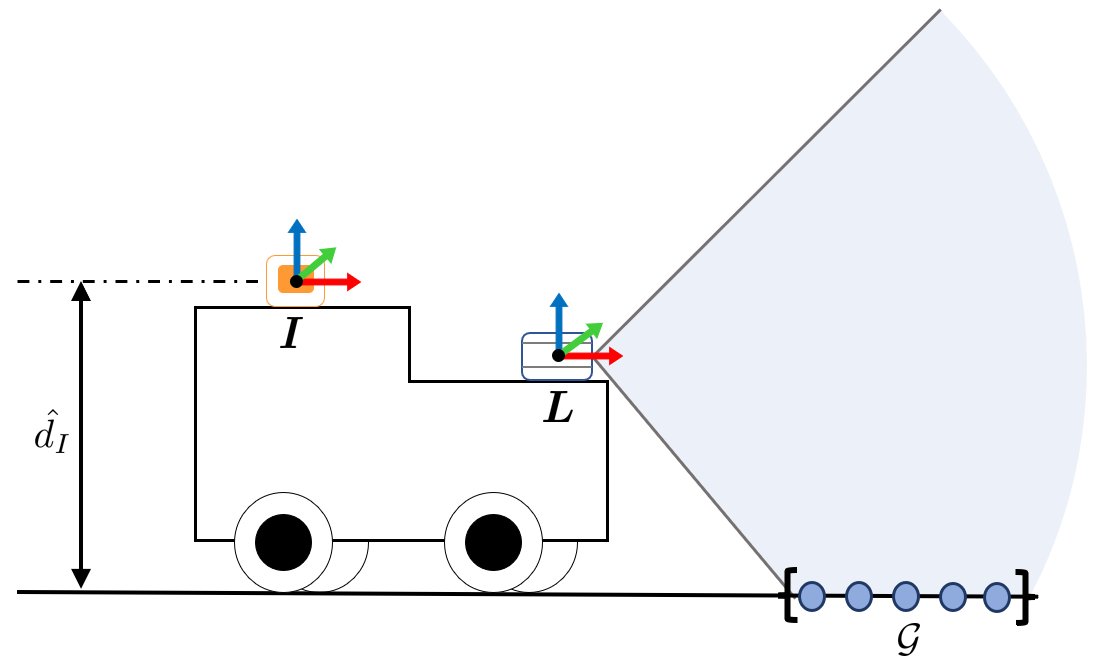}
\caption{ Illustration of a ground robot equipped IMU ($I$) sensor and LiDAR ($L$) sensor. We leverage the IMU height from the ground  ($\hat{d}_I$) as minimal prior knowledge and the LiDAR ground segmentation points ($\mathcal{G}$) for IMU-LiDAR extrinsic calibration.}
\label{fig1}
\end{figure}

Therefore, a lot of research has been conducted regarding the extrinsic calibration between IMU and LiDAR, and the research can be divided into two directions: target-based and targetless approaches. Target-based approach uses known, predefined objects such as cylinders \cite{liu2019error} to estimate the spatial transformations between IMU and LiDAR. Target-based approach is relatively accurate, but it has the shortcoming that predetermined targets are required. On the other hand, the targetless approach does not use any predefined objects for calibration, receiving much attention due to ease of use and repeatability. Instead, the targetless method requires comprehensive excitation motion of both sensors to activate all Degrees of Freedom (DoF). Achieving this motion is straightforward for handheld devices with integrated IMU and LiDAR sensors, as they can easily perform a full range of motions. However, for ground robots, it becomes a significant challenge because their motion is limited to planar motion. Furthermore, from a theoretical perspective, the lack of full motion will make the rank of Fisher information matrix (FIM) less than its dimension, and the degenerate or rank-deficient FIM will cause the estimated calibration parameters to have high uncertainty or not to become unique. To address the issue, some papers perform observability analysis \cite{yang2019degenerate}, \cite{yang2023online}, \cite{lv2022observability}, but the analysis only identifies unobservable directions and does not directly estimate calibration parameters. 

In this paper, we present \textit{GRIL-Calib}: novel targetless Ground Robot IMU-LiDAR extrinsic Calibration method. To address the aforementioned degenerate problem, we utilize a ground plane motion (GPM) constraint. The GPM provides geometric information about the robot's motion and allows the full 6-DoF extrinsic parameters even when some of the extrinsic parameters are unobservable. The extrinsic calibration problem between IMU and LiDAR for ground robots is depicted in Fig \ref{fig1}. Similar constraints related to the ground have been proposed for LiDAR odometry (LO) \cite{wei2022gclo} or visual-inertial odometry (VIO) \cite{li2022visual}. However, we propose a ground plane motion (GPM) constraint specifically focused on the extrinsic calibration of the IMU and LiDAR. The proposed method leverages a motion-based approach and integrates ground observation, thereby offering a user-friendly and accurate solution for extrinsic calibration of the IMU and LiDAR mounted on the ground robot. The key contributions of this paper are summarized as follows:

\begin{itemize}
 
\item We propose a novel extrinsic calibration method of IMU-LiDAR fusion system on a ground robot and thus does not need to satisfy full excitation motion. 

\item To improve the performance of the calibration, we proposed a novel LiDAR Odometry (LO) that exploits ground plane observations. The ground plane residuals are incorporated into an Iterative Error State Kalman Filter (IESKF) to reduce drift around the z-axis.

\item We propose Ground Plane Motion (GPM) constraints and apply the constraint to the nonlinear optimization for calibration. The GPM enables us to address parameters including unobservable direction and also improve calibration accuracy. 

\item In the nonlinear optimization for calibration, we formulate the calibration of the translation and rotation parameters as a Single Optimization (SO) problem solve parameters simultaneously. The baseline methods, however, calibrate the two parameters separately and sequentially, degrading the calibration accuracy.  

\item We validated \textit{GRIL-Calib} on a variety of datasets, achieving significant results compared to state-of-the-art targetless methods. We have open-sourced\footnote{\url{https://github.com/Taeyoung96/GRIL-Calib}} our algorithm to benefit the broader robotics community.

\end{itemize}

The rest part of this paper is organized as follows: Section \ref{related-works} presents a comprehensive review of recent IMU-LiDAR calibration methods. In Section \ref{overview}, we present the system overview of \textit{GRIL-Calib} and describe each key part in detail at Section \ref{preprocessing} and Section \ref{optimization}. Section \ref{experiments} describes the experiments performed on various real-world datasets and the analysis of the results. Finally, Section \ref{conclusions} provides conclusions and discusses future work.

\section{Related Work}
\label{related-works}

In early IMU-LiDAR calibration research, an arbitrary target is required for calibration \cite{liu2019error}, \cite{8460179}. However, the target-based method is not practical, so targetless calibration methods have been studied more recently. Subodh \textit{et al}. \cite{mishra2021target} formulate a motion-based calibration based on the Extended Kalman Filter (EKF). Additionally, Lv \textit{et al}. \cite{lv2020targetless} utilizes a continuous-time trajectory formulation based on B-Spline. Also, OA-LICalib \cite{lv2022observability} which extends their previous works \cite{lv2020targetless}, proposes two observability-aware modules designed to address the degenerate case. Unfortunately, the proposed observability-aware module is designed to only reject unnecessary motion. FAST-LIO2 \cite{xu2022fast} is an accurate LiDAR-inertial odometry (LIO) algorithm based on an Iterated error state Kalman filter (IESKF) which include extrinsic calibration parameters. Due to the strong nonlinearity, the performance of the LIO system is highly dependent on accurate initial states. To get an accurate estimate of the initial states, Zhu \textit{et al}. \cite{zhu2022robust} propose the optimization-based algorithm that aligns angular velocity and acceleration. Existing approaches \cite{mishra2021target}, \cite{lv2022observability}, \cite{zhu2022robust} require sufficient motion in all axes of the sensor to obtain accurate results. These methods are practical when applied to handheld devices that can move freely in various directions. However, existing methods encounter challenges when applied to ground robots where motion is limited to planar motion. In contrast, our method utilizes ground observation that can easily be observed by a ground robot to obtain accurate 6-DoF extrinsic parameters, even in the absence of sufficient motion in all axes. Recently, there are also several approaches to calibrating different sensors on ground robots. Zuniga-Noel \textit{et al}. \cite{8735740} proposed a two-step multi-sensor calibration method for mobile robots using a camera and 2D laser. Also, an IMU-LiDAR calibration algorithm utilizing a GNSS sensor in planar motion has been proposed \cite{10106780} and \cite{10011214}. However, they require an additional GNSS sensor, while our proposed algorithm calibrates IMU-LiDAR without additional sensors.

\section{System Overview}
\label{overview}

\begin{figure}[t]
\centering
\includegraphics[width=0.8\columnwidth]{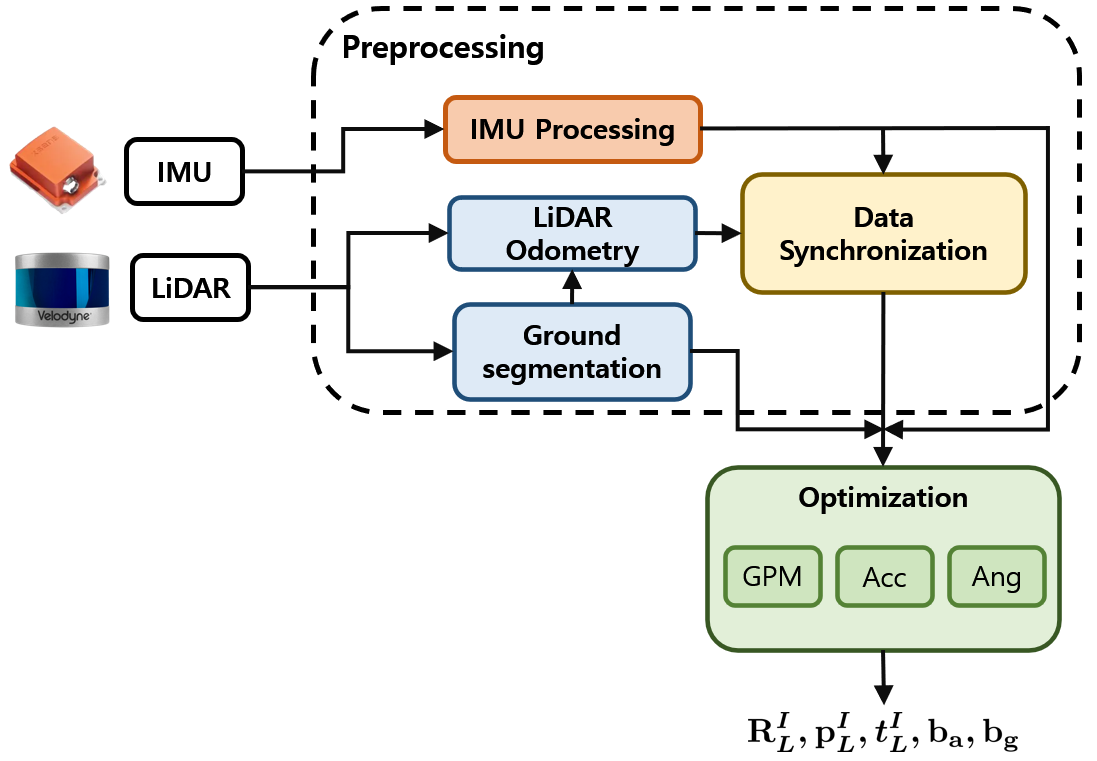}

\caption{The overview of proposed \textit{GRIL-Calib} system. During the optimization stage (Green Box), we use 3 constraints. See  Section \ref{optimization} for more details.}
\label{fig2}
\end{figure}

The overall system of \textit{GRIL-Calib} comprises two primary stages: pre-processing and optimization, as shown in Fig. \ref{fig2}. Our framework is motivated by \cite{zhu2022robust}. The first stage is pre-processing, which generates additional information that is critical for the following optimization stage. Specifically, the pre-processing stage performs the following key tasks: 
\begin{itemize}

\item \textit{Ground Observation via ground segmentation}, where ground points are extracted from LiDAR data, facilitating the estimation of the LiDAR sensor's relative orientation to the ground.

\item \textit{Lidar Odometry (LO)}, which enhances LO performance by utilizing the extracted ground plane and incorporating a proposed ground plane residual.

\item \textit{IMU Processing}, involving the pre-processing of IMU measurements and the calculation of the IMU's relative orientation to the ground using the Madgwick filter \cite{madgwick2010efficient}. 

\item \textit{Data Synchronization}, aimed at determining the approximate time offset between IMU and LiDAR odometry using cross-correlation methods that analyze their magnitude of angular velocity. Notably, this process aligns with the approach proposed in \cite{zhu2022robust}.

\end{itemize}

The second stage is optimization, which aims to optimize the extrinsic calibration parameters by utilizing the data acquired in the preprocessing stage. In this optimization stage, we apply three types of constraints: unified temporal-spatial constraints based on angular velocity and linear acceleration, as well as proposed Ground Plane Motion (GPM) constraints. The GPM constraint ensures that the robot's motion remains on the ground and is integrated into the objective function to achieve precise extrinsic calibration while limiting planar motion. In contrast \cite{zhu2022robust}, where each type of objective function is optimized individually and sequentially, we simultaneously optimize all three types of constraints by assigning different values of weights to each objective. This approach improves performance compared to the method in \cite{zhu2022robust}.

Throughout the paper, we use the following notations: LiDAR frame denotes $\left\{ L \right\}$, IMU frame denotes $\left\{ I \right\}$, $\mathbf{R}^I_L \in SO(3)$ is the rotation matrix from IMU frame to LiDAR frame. $\mathbf{p}^I_L \in \mathbb{R}^3 $ is the translation vector from the IMU frame to the LiDAR frame. $t^I_L \in \mathbb{R}$ is the time offset between IMU and LiDAR. $\mathbf{b}_a, \mathbf{b}_g \in \mathbb{R}^3$ means the acceleration bias and angular velocity bias of the IMU, respectively.

\section{Preprocessing}
\label{preprocessing}

In the preprocessing stage, we obtain some parameters and the sensor measurements which will be used in the subsequent the optimization stage. 

\subsection{Ground Observation via LiDAR Ground Segmentation}

First, the ground is segmented from the LiDAR input. LiDAR ground segmentation aims to obtain the rotation matrix $\mathbf{R}^G_{L}$ of the LiDAR with respect to the ground. When the current LiDAR scan $\mathcal{P}$ is presented, we extract a set of ground points $\mathcal{G}$ using Patchwork++ \cite{lee2022patchwork++}, a state-of-the-art (SOTA) ground segmentation algorithm. Then, we apply the Principal Component Analysis (PCA) to the covariance matrix $\mathcal{C}_{\mathcal{G}}$ of the ground points $\mathcal{G}$ identifying a normal vector $\mathbf{n}_L  = (a_L,b_L,c_L)^T$ of the ground plane 
\begin{equation}
a_L x + b_L y + c_L z + d_L = 0
\label{plane-eq}
\end{equation}
 where $\mathbf{n}_L$ implies a normal vector to the ground in the LiDAR coordinate $\left\{ L \right\}$. We also define a normal vector $\mathbf{n}_G  = (0, 0, 1)^T$ with respect to the ground coordinate $\left\{ G \right\}$. From the two normal vectors $\mathbf{n}_L$  and $\mathbf{n}_G$ , we obtain a rotation matrix $\mathbf{R}^G_{L}$  from  $\left\{ G \right\}$ frame  to $\left\{ L \right\}$ frame  by 
\begin{equation}
\mathbf{R}_{L}^{G}=\exp ({{\left[ {{\phi }_{GL}}{{\mathbf{n}}_{GL}} \right]}_{\times }})     
\label{R_ground2lidar}
\end{equation}
where ${{\phi }_{GL}}=\arccos ({{\mathbf{n}}_{G}}^{T}{{\mathbf{n}}_{L}})\in \mathbb{R}$ and ${{\mathbf{n}}_{GL}}={{\mathbf{n}}_{G}}\times {{\mathbf{n}}_{L}}\in {{\mathbb{R}}^{3}}$ denote the rotation angle and vector between  $\left\{ G \right\}$ and  $\left\{ L \right\}$, respectively. $\left[ \cdot \right]_{\times}$ denotes the skew-symmetric matrix, and $\exp \left( \cdot \right)$  denotes the exponential map.

\subsection{LiDAR Odometry Utilizing the Ground Plane Residual}

We use Iterated Error state Kalman Filter (IESKF) to obtain LiDAR Odometry (LO) with a similar procedure to \cite{zhu2022robust}. However, we improved the performance of LO by exploiting the assumption that \textit{ground robot is constrained to movement on a 2D plane.} When we define the state vector $\mathbf{x_k}$ of the LiDAR by  
\begin{equation}
{{\mathbf{x}}_{k}}={{\left[ \log\left( {{\mathbf{R}}_{{{L}_{k}}}} \right),{{\mathbf{p}}_{{{L}_{k}}}},{{\boldsymbol{\omega }}_{{{L}_{k}}}},{{\mathbf{v}}_{{{L}_{k}}}} \right]}^{T}}
\end{equation}
and its motion as follows,
\begin{equation}
\mathbf{x}_{k+1}=\mathbf{x}_k \boxplus\left(\Delta t \cdot \mathbf{F}\left(\mathbf{x}_k, \mathbf{w}_k\right)\right)
\end{equation}
where $\mathbf{F}({{\mathbf{x}}_{\mathbf{k}}},{{\mathbf{w}}_{\mathbf{k}}})={{\left[ {{\boldsymbol{\omega }}_{{{L}_{k}}}},{{\mathbf{v}}_{{{L}_{k}}}},\boldsymbol{{{\varepsilon }}_{\boldsymbol{\omega }}},{\boldsymbol{{\varepsilon}}_{\mathbf{v}}} \right]}^{T}}$. $\mathbf{R}_{L_k}$ represents the attitude from the initial LiDAR frame $\left\{ {{L}_{0}} \right\}$ to the LiDAR frame $\left\{ {{L}_{k}} \right\}$ at current time $k$, that is ${{\mathbf{R}}_{{{L}_{k}}}}=\mathbf{R}_{{{L}_{k}}}^{{{L}_{0}}}$, $\log \left( \cdot \right)$ denotes the logarithmic map. ${{\mathbf{p}}_{{{L}_{k}}}}=\mathbf{p}_{{{L}_{k}}}^{{{L}_{0}}}$ represents the position of the LiDAR at current time $k$.  ${{\boldsymbol{\omega }}_{{{L}_{k}}}}$ and $\mathbf{v}_{L_k}$ represent the angular and linear velocities of the LiDAR at time $k$, respectively. ${{\mathbf{w}}_{k}}=\left[ {\boldsymbol{{\varepsilon}}_{\boldsymbol{\omega }}},{\boldsymbol{{\varepsilon}}_{\mathbf{v}}} \right]\sim\mathcal{N}\left( \mathbf{0},{{\mathbf{Q}}_{k}} \right)$ are a random walk noise represented by zero-mean Gaussian.  $\Delta t$ is the time interval between two consecutive scans, and the notation $\boxplus$ represents the encapsulated ``boxplus" operations on the state manifold \cite{hertzberg2013integrating}.

The prediction step of the IESKF consists of the prediction of the state and the propagation of the covariance as follows,

\begin{equation}
\begin{aligned}
\hat{\mathbf{x}}_{k+1}&=\bar{\mathbf{x}}_k \boxplus\left(\Delta t \cdot \mathbf{F}\left(\bar{\mathbf{x}}_k, \mathbf{0}\right)\right) \\
\hat{\mathbf{P}}_{k+1}&=\mathbf{F}_{\tilde{\mathbf{x}}} \bar{\mathbf{P}}_k \mathbf{F}_{\tilde{\mathbf{x}}}^T+\mathbf{F}_{\mathbf{w}} \mathbf{Q}_k \mathbf{F}_{\mathbf{w}}^T
\end{aligned}
\end{equation}
where $\hat{(\cdot)}$, $\tilde{(\cdot)}$, and $\bar{(\cdot)}$ denote the predicted, error and optimal states, respectively; $\mathbf{P}_k$ and $\mathbf{Q}_k$ are the covariance matrices for error state $\tilde{\mathbf{x}}_k$  and noise $\mathbf{w}_k$, respectilvey. $\mathbf{F}_{\tilde{\mathbf{x}}}$ and $\mathbf{F}_{\mathbf{w}}$ are the derivative of $(\mathbf{x}_{k+1} \boxminus \hat{\mathbf{x}}_{k+1})$ with respect to $\tilde{\mathbf{x}}$ and $\mathbf{w}$ evaluated at zero, respectively. The notation $\boxminus$ represents the encapsulated ``boxminus" operations on the state manifold \cite{hertzberg2013integrating}.

The correction step of IESKF is conducted by a Maximum A Posteriori (MAP) problem as in \cite{zhu2022robust}, where the prior distribution is obtained through state prediction, and the likelihood is obtained through the first-order approximation of point-to-plane measurement model \cite{xu2022fast}. 

$\mathcal{P}{{}_{k}}=\left\{ p_{j}^{{{L}_{k}}} \;|\;j=1,\cdots,{{N}_{k}} \right\}$ is given at time $k$, the measurement model \cite{xu2022fast} is constructed based on the idea that a current point $p_{j}^{{{L}_{k}}}$ acquired at time $k$ should remain on a corresponding  neighboring plane ${{\pi }_{j}}$ in the map, as described below:
\begin{equation}
0={{\mathbf{h}}_{j}}\left( {{\mathbf{x}}_{k}},p_{j}^{{{L}_{k}}}+\boldsymbol{\varepsilon }_{j}^{{{L}_{k}}} \right)={{\left( \mathbf{u}_{j}^{{{L}_{0}}} \right)}^{T}}\left( \mathbf{T}_{{{L}_{k}}}^{{{L}_{0}}}\left( p_{j}^{{{L}_{k}}}+\boldsymbol{\varepsilon }_{j}^{{{L}_{k}}} \right)-\mathbf{q}_{j}^{{{L}_{0}}} \right)
\label{point2plane}
\end{equation}
where $\mathbf{u}_{j}^{{{L}_{0}}}$ is the unit normal vector of the nearest corresponding plane defined w.r.t $\left\{ {{L}_{0}} \right\}$; $\mathbf{T}_{{{L}_{k}}}^{{{L}_{0}}}$ represent a LiDAR pose including ($\mathbf{R}_{L_k}$,$\mathbf{p}_{L_k}$), and the point $\mathbf{q}_{j}^{{{L}_{0}}}$ is a LiDAR point located on the plane within the map. By applying the first-order approximation to (\ref{point2plane}), we obtain
\begin{equation}
\begin{aligned}
0&={{\mathbf{h}}_{j}}\left( {{\mathbf{x}}_{k}}, \; p_{j}^{{{L}_{k}}}+\boldsymbol{\varepsilon }_{j}^{{{L}_{k}}} \right) \\
&\simeq {{\mathbf{h}}_{j}}\left( \hat{\mathbf{x}}_{k}^{{}}, \; p_{j}^{{{L}_{k}}}+0 \right)+\mathbf{H}_{j}^{{}}\tilde{\mathbf{x}}_{k}^{{}}+{{\mathbf{v}}_{j}}=\mathbf{z}_{j}^{{}}+\mathbf{H}_{j}^{{}}\tilde{\mathbf{x}}_{k}^{{}}+{{\mathbf{v}}_{j}}    
\end{aligned}
\label{h_j}
\end{equation}
where  $\mathbf{z}_{j}^{{}}={{\mathbf{h}}_{j}}\left( \hat{\mathbf{x}}_{k}^{{}},p_{j}^{{{L}_{k}}}+0 \right)$, $\mathbf{H}_{j}^{{}}={{\left. \frac{\partial {{\mathbf{h}}_{j}}}{\partial {{\mathbf{x}}_{k}}} \right|}_{{{\mathbf{x}}_{k}}={{{\mathbf{\tilde{x}}}}_{k}}}}$, and \\
${{\mathbf{v}}_{j}} \in \mathcal{N}\left( \mathbf{0},{{\mathbf{R}}_{j}} \right)$ is the disturbance which comes from the raw measurement noise $\boldsymbol{\varepsilon }_{j}^{{{L}_{k}}}$. 

Different from \cite{zhu2022robust}, however, we enhance the LO by \textit{exploiting the planar constraint} which effectively reduces drift in the LO. Specifically, we further use the constraint that the roll and pitch of the rotation matrix ${{\mathbf{R}}_{{{L}_{k}}}}$ and the z-component of the translation vector $\mathbf{p}_{L_k}$ should be consistent because the motion of the robot is restricted to a 2D space. Thus, when we are given a normal vector ${{\mathbf{n}}_{{{L}_{k}}}}$ of the LiDAR ground points ${{\mathcal{G}}_{k}}$ at time $k$, the vector ${{\mathbf{n}}_{{{L}_{k}}}}$ should be orthogonal to the x-y plane of the ground frame $\left\{ G \right\}$. 
\begin{equation}
{{[\mathbf{R}_{L}^{G}{{\mathbf{R}}_{{{L}_{k}}}}({{\mathbf{n}}_{{{L}_{k}}}}+{{\boldsymbol{\varepsilon }}_{{{\mathbf{n}}_{k}}}})]}_{1,2}}=0
\label{gpm1}
\end{equation}
where ${{\boldsymbol{\varepsilon }}_{{{\mathbf{n}}_{k}}}}$ is the observation noise to ${{\mathbf{n}}_{{{L}_{k}}}}$. $ \lbrack \cdot \rbrack_{1,2}$ denotes the notation that takes only the first and second elements in a ($3 \times 1$) vector is used. Further, the displacement of the z-component of ${{\mathbf{p}}_{{{L}_{k}}}}$ should be zero respect to ground frame $\left\{ G \right\}$, that is,
\begin{equation}
\mathbf{e}_{3}^{T}\mathbf{R}{{_{L}^{G}}^{{}}}{{\mathbf{p}}_{{{L}_{k}}}}=0.     
\label{gpm2}
\end{equation}
By combining (\ref{gpm1}) and (\ref{gpm2}), we define a ground plane measurement model $\mathbf{h}_{\pi}$ given by
\begin{equation}
0={{\mathbf{h}}_{\pi }}\left( {{\mathbf{x}}_{k}},{{\mathbf{n}}_{{{L}_{k}}}}+{{\boldsymbol{\varepsilon}}_{{{\mathbf{n}}_{k}}}} \right)=\left( \begin{matrix}
   {{\left[ \mathbf{R}_{L}^{G}{{\mathbf{R}}_{{{L}_{k}}}}\left( {{\mathbf{n}}_{{{L}_{k}}}}+{{\mathbf{\varepsilon }}_{{{\mathbf{n}}_{k}}}} \right) \right]}_{1,2}}  \\
   \mathbf{e}_{3}^{T}\mathbf{R}_{L}^{G}{{\mathbf{p}}_{{{L}_{k}}}}  \\
\end{matrix} \right)
\label{gpm}
\end{equation}

Using (\ref{gpm}), by applying the first-order approximation, we derive the following equation,
\begin{equation}
\begin{aligned}
0=& \;{{\mathbf{h}}_{\pi }}\left( {{\mathbf{x}}_{k}},{{\mathbf{n}}_{{{L}_{k}}}}+{{\boldsymbol{\varepsilon}}_{{{\mathbf{n}}_{k}}}} \right) \\
\simeq& \; {{\mathbf{h}}_{\pi }}\left( \hat{\mathbf{x}}_{k}^{{}},{{\mathbf{n}}_{{{L}_{k}}}}+\mathbf{0} \right)+\mathbf{H}_{\pi }^{{}}\tilde{\mathbf{x}}_{k}^{{}}+{{\mathbf{v}}_{\pi }}=\mathbf{z}_{\pi }^{{}}+\mathbf{H}_{\pi }^{{}}\tilde{\mathbf{x}}_{k}^{{}}+{{\mathbf{v}}_{\pi }} 
\end{aligned}
\end{equation}
where $\mathbf{z}_{\pi }^{{}}={{\mathbf{h}}_{\pi }}\left( \hat{\mathbf{x}}_{k}^{{}},{{\mathbf{n}}_{{{L}_{k}}}}+\mathbf{0} \right)$, $\mathbf{H}_{\pi }^{{}}={{\left. \frac{\partial {{\mathbf{h}}_{\pi }}}{\partial {{\mathbf{x}}_{k}}} \right|}_{{{\mathbf{x}}_{k}}={{{\mathbf{\tilde{x}}}}_{k}}}}$ \\ ,and  $\mathbf{v_{\pi}} \sim \mathcal{N}\left(\mathbf{0}, \mathbf{S}_{k}\right)$.

By integrating the two likelihoods given by $\mathbf{h}_j$ in (\ref{h_j}) and our proposed $\mathbf{h}_{\pi}$ in (\ref{gpm}) apply to MAP estimate, we obtain ${{\tilde{x}}_{k}}$ by
\begin{equation}
\min _{\tilde{\mathbf{x}}_k^\tau}\left(\left\|\mathbf{x}_k \boxminus \hat{\mathbf{x}}_k\right\|_{\hat{\mathbf{P}}_k}^2+\sum_{j=1}^m\left\|\mathbf{z}_j^\tau+\mathbf{H}_j^\tau \tilde{\mathbf{x}}_k^\tau\right\|_{\mathbf{R}_j}^2 + \left\|\mathbf{z}_{\pi}^\tau+\mathbf{H}_{\pi}^\tau \tilde{\mathbf{x}}_k^\tau\right\|_{\mathbf{S}_k}^2\right ) 
\end{equation}
where $\tau$ means the number of iterate times. $m$ is the number of LiDAR points.
When $\tilde{\mathbf{x}}_k$ decreases below a certain threshold, the state is considered to be optimal.
Based on the optimal state $\bar{\mathbf{x}}_k$ we return,
\begin{equation}
\mathbb{L}_k=\left\{
\boldsymbol{\omega}_{L_k}, 
\mathbf{v}_{L_k},
\boldsymbol{\alpha}_{L_k},
\mathbf{a}_{L_k}
\right\}
\label{lo_values}
\end{equation}
In (\ref{lo_values}), ${{\boldsymbol{\omega }}_{{{L}_{k}}}}$ and ${{\mathbf{v}}_{{{L}_{k}}}}$ are obtained from $\bar{\mathbf{x}}_k$, and the angular acceleration $\boldsymbol{\alpha}_{L_k}$ and the linear acceleration $\mathbf{a}_{L_k}$ of the LiDAR are computed by applying non-causal central difference \cite{baleanu2009central} to ${{\boldsymbol{\omega }}_{{{L}_{k}}}}$ and ${{\mathbf{v}}_{{{L}_{k}}}}$, respectively.  

\begin{figure}[t]
\centering
\includegraphics[width=0.95\columnwidth]{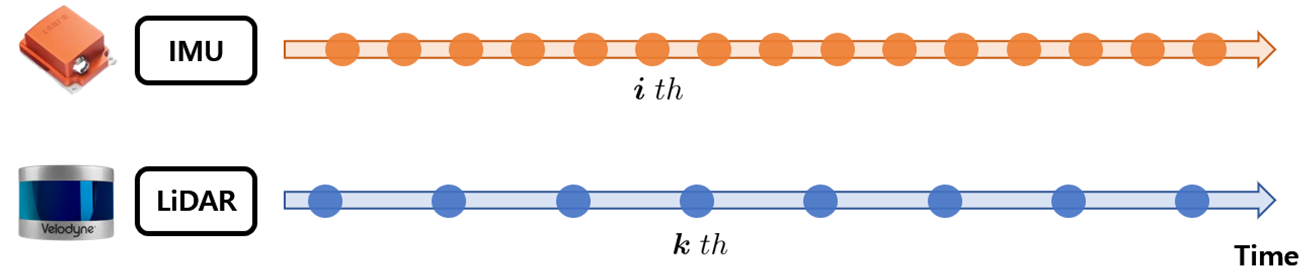}
\caption{Visualization of timestamp for each IMU measurement and LiDAR Odometry.}
\label{downsample}
\end{figure}

\subsection{IMU Processing and Synchronization with LiDAR Odometry}
In general, the IMU measurement model is defined as follows.  
\begin{equation}
\hat{\boldsymbol{\omega}}_i=\boldsymbol{\omega}_i+\mathbf{b}_g+\boldsymbol{\varepsilon}_g, \quad \hat{\mathbf{a}}_i=\mathbf{a}_i+ \mathbf{R}^w_{I_i} \mathbf{g}^w +\mathbf{b}_a+\boldsymbol{\varepsilon}_a 
\label{imu}
\end{equation}
The raw gyroscope and acceleration measurements at time $i$ denote as $\hat{\boldsymbol{\omega}}_i$ and $\hat{\mathbf{a}}_i$. $\mathbf{b}_g$ is the gyroscope bias and $\boldsymbol{\varepsilon}_g$ is the gaussian white noise for the gyroscope measurement. Also, $\mathbf{b}_a$ is the acceleration bias and $\mathbf{g}^w$ is the gravity vector relative to world frame $w$, and $\boldsymbol{\varepsilon}_a$ is a Gaussian white noise for the acceleration measurement. Here, it should be noted that the IMU measurements are much faster than LO, we use two different time indices $i$ and $k$ for IMU and LO, respectively, as shown in Fig \ref{downsample}. 

Similar to \cite{zhu2022robust}, we remove the noise $\boldsymbol{\varepsilon}_g$ and $\boldsymbol{\varepsilon}_a$ from the IMU measurements with a zero-phase filter \cite{gustafsson1996determining} in (\ref{imu}). 
Also, we down-sample to get the IMU measurement at timestamp $k$ from the LiDAR. 
\begin{equation}
\boldsymbol{\bar{\omega}}_{I_k} = \boldsymbol{\omega}_{I_k}+\mathbf{b}_g,
    \quad \bar{\mathbf{a}}_{I_k}=\mathbf{a}_{I_k}+ \mathbf{R}^w_{I_k} \cdot \mathbf{g}^w +\mathbf{b}_a
\end{equation}

Since the IMU and LiDAR Odometry are not synchronized, we utilize the cross-correlation between their respective angular velocities to quantify the similarity between their magnitudes. To achieve more precise synchronization between two sensors, we employ a coarse-to-fine approach to determine the time offset, denoted as $t^I_L$. As outlined in \cite{zhu2022robust}, $t^I_L$ is the sum of the coarse estimate $t_c$ and the fine estimate $t_f$, thereby defined as $t^I_L = t_c + t_f$. Similar to \cite{zhu2022robust}, we determine $t_c$ by using the cross-correlation method given by   
\begin{equation}
t_c^*=\underset{t_c}{\arg \max } \sum_{k=1}^n\left\| \bar{\boldsymbol{\omega}}_{I_{k+{t}_c}}\right\| \cdot\left\|\boldsymbol{\omega}_{L_k}\right\|  
\label{cross_corr}
\end{equation}
In the following section, we derive the remaining component, $t_f$, as the outcome of the optimization process. Further, by applying the IMU measurements ${{\boldsymbol{\omega }}_{{{I}_{k}}}}$ and ${{\mathbf{a}}_{{{I}_{k}}}}$ to the Madgwick filter \cite{madgwick2010efficient}, we estimate the rotation matrix $\mathbf{R}^G_I$ from the ground frame $\left\{ G \right\}$ to the IMU frame $\left\{ I \right\}$. Finally, as a result from the IMU processing and synchronization procedure, we return 
\begin{equation}
{{\mathbb{I}}_{k}}=\left\{ {{{\boldsymbol{\bar{\omega }}}}_{{{I}_{k}}}},{{{\bar{\boldsymbol{\alpha }}}}_{{{I}_{k}}}},{{{\mathbf{\bar{a}}}}_{{{I}_{k}}}},\mathbf{R}_{I}^{G} \right\}
\label{imu_meas}
\end{equation}

\section{EXTRINSIC CALIBRATION VIA SINGLE OPTIMIZATION (SO)}
\label{optimization}
Using the two sets of measurements ${{\mathbb{L}}_{k}}$ in (\ref{lo_values}) and ${{\mathbb{I}}_{k}}$ in (\ref{imu_meas}) obtained from Section \ref{preprocessing}, we formulate the IMU-LiDAR extrinsic calibration as an optimization problem. The calibration parameter $\mathbf{x}_{cal}$ used in the calibration is given by,
\begin{equation}
\mathbf{x}_{cal}=\left[
\mathbf{R}^I_L, 
\mathbf{p}^I_L, 
t_f, 
\mathbf{b}_g,
\mathbf{b}_a
\right]
\end{equation}

\subsection{Unified Temporal-Spatial Constraints}
We jointly utilize the two constraints proposed in \cite{zhu2022robust}. 
The first constraint uses the angular velocity measurements ${{\bar{\mathbf{\omega}}}_{{{I}_{k}}}}$ and ${{\mathbf{\omega }}_{{{L}_{k}}}}$ from IMU measurements and LO, respectively. 
\begin{equation}
\bar{\boldsymbol{\omega}}_{I_{k^{\prime}}} + t_f \cdot  \boldsymbol{\bar{\alpha}}_{I_{k^{\prime}}}= (\mathbf{R}^I_L)^T \boldsymbol{\omega}_{L_k} +\mathbf{b}_g 
\label{angular-eqation}
\end{equation}
by compensating the time offset $k^{\prime}=k+t_c^*$ between two sensors, where $t_c^*$ is given in (\ref{cross_corr}) and assuming the IMU angular acceleration $\boldsymbol{\bar{\alpha}}_{I_{k^{\prime}}}$ is constant over a small period of time $t_f$. This is described as follows, according to (\ref{angular-eqation})
\begin{equation}
\mathbf{r}_w (\mathbf{R}^I_L, \mathbf{b}_g, \mathbf{t}_f ) = (\mathbf{R}^I_L)^\mathbf{T}  \boldsymbol{\omega}_{L_k} +\mathbf{b}_g - \bar{\boldsymbol{\omega}}_{I_{k^{\prime}}}- t_f \cdot  \boldsymbol{\bar{\alpha}}_{I_{k^{\prime}}}  \end{equation}

The second constraint uses the acceleration measurements ${{{\mathbf{\bar{a}}}}_{{{I}_{k}}}}$ and ${{\mathbf{a}}_{{{L}_{k}}}}$ from IMU and LO, respectively. As in \cite{xu2022robots}, the accelerations ${{\mathbf{a}}_{{{I}_{k}}}}$ and ${{\mathbf{a}}_{{{L}_{k}}}}$of the two sensors are related to each other by
\begin{equation}
\mathbf{R}^I_L {\mathbf{\bar{a}}}_{I} = \mathbf{a}_{L} 
+ \{([ \boldsymbol{\omega}_{L}]_{\times}^2 + [\boldsymbol{\alpha}_{L}]_{\times}) \cdot \mathbf{p}^I_L \}
\end{equation}
where $\boldsymbol{\alpha}_{L}$ is obtained from ${{\mathbb{L}}_{k}}$. the second constraint $\mathbf{r}_a$ is represented as follows. 
\begin{equation}
\begin{aligned}
\mathbf{r}_a (\mathbf{R}^I_L,\mathbf{p}^I_L, \mathbf{b}_a) &= \mathbf{R}^I_L(\mathbf{\bar{a}}_{I_{k^{\prime}}}- \mathbf{b}_a) - \mathbf{a}_{L_k} \\
& \quad- \{([ \boldsymbol{\omega}_{L_k}]_{\times}^2 + [\boldsymbol{\alpha}_{L_k}]_{\times}) \cdot \mathbf{p}^I_L \}
\end{aligned}
\end{equation}

\subsection{Ground Plane Motion Constraints}
\begin{figure}[t]
\centering
\includegraphics[width=0.65\columnwidth]{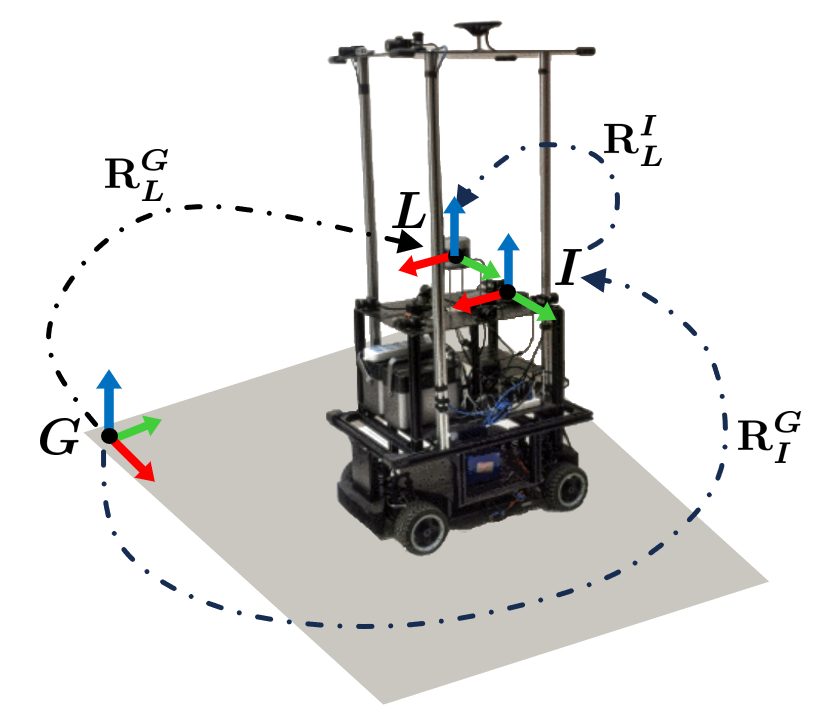}

\caption{Illustration of the geometric rotation relationship between ground and each sensor. This robot platform is used in the M2DGR dataset.
}
\label{ground-robot}
\end{figure}

In this subsection, we describe a novel constraint named Ground Plane Motion (GPM). Our proposed constraint is slightly similar to the constraint proposed in \cite{li2022visual}, where the plane motion information was leveraged to improve the performance of VINS. However, the constraint in \cite{li2022visual} was formulated for the estimation of sensor motion, while our constraint is specifically designed for extrinsic calibration between IMU and LiDAR. Specifically, our GPM is parameterized in terms of $\mathbf{R}_{L}^{I}$ and $\mathbf{p}_{L}^{I}$. In the formulation of the GPM, we directly measure the height of the IMU from the ground $\hat{d_I} \in \mathbb{R}$ and use it as prior knowledge. Further, we exploit the fact that the z-component of $\mathbf{p}{{_{L}^{I}}_{{}}}$ could not be observed because it is in the direction perpendicular to the planar motion. Using the geometric relationship of the rotation parameters between the ground and each sensor shown in Fig \ref{ground-robot}, we obtain 
\begin{equation}
{{[\mathbf{R}_{I}^{G}\mathbf{R}_{L}^{I}\mathbf{R}_{G}^{L}\mathbf{e}_{3}^{{}}]}_{1,2}}=0
\end{equation}
\begin{equation}
\mathbf{e}_{3}^{T}{{(\mathbf{R}_{I}^{L}\mathbf{R}_{G}^{I}{{\mathbf{d}}_{I}}-\mathbf{R}_{G}^{L}{{\mathbf{d}}_{L}})}_{{}}}=\mathbf{e}_{3}^{T}\mathbf{p}{{_{L}^{I}}_{{}}}
\end{equation}
where ${{\mathbf{d}}_{L}}={{d}_{L}}\mathbf{e}_{3}^{{}}$, and ${{d}_{L}}$ is the height of the LiDAR from the ground and it is obtained from (\ref{plane-eq}). ${{\mathbf{d}}_{I}}=\hat{{{d}_{I}}}\mathbf{e}_{3}^{{}}$, and $\hat{{{d}_{I}}}$ denotes the height of the IMU from the ground
${{(\mathbf{R}_{L}^{G})}^{T}}=\mathbf{R}_{G}^{L}$ and ${{(\mathbf{R}_{I}^{G})}^{T}}=\mathbf{R}_{G}^{I}$, and they are obtained from (\ref{R_ground2lidar}) and (\ref{imu_meas}), respectively. The proposed Ground Plane Motion (GPM) constraint  $\mathbf{r}_g$  is represented as
\begin{equation}
{{\mathbf{r}}_{g}}(\mathbf{R}_{L}^{I},\mathbf{p}_{L}^{I})=\left( \begin{matrix}
{{[\mathbf{R}_{I}^{G}\mathbf{R}_{L}^{I}\mathbf{R}_{G}^{L}\mathbf{e}_{3}^{{}}]}_{1,2}}  \\
\mathbf{e}_{3}^{T}{{(\mathbf{R}_{I}^{L}\mathbf{R}_{G}^{I}{{\mathbf{d}}_{I}}-\mathbf{R}_{G}^{L}{{\mathbf{d}}_{L}}-\mathbf{p}_{L}^{I})}_{{}}}  \\
\end{matrix} \right)
\end{equation}

Unlike \cite{10106780}, which performs the calibration while ignoring unobservable direction, our \textit{Gril-Calib} conducts extrinsic calibration including unobservable direction even under the planar motion by utilizing Ground Plane Motion (GPM) constraints.

\subsection{Single Optimization (SO) Process}
The method proposed in \cite{zhu2022robust} uses two distinct constraints, $\mathbf{r}_w$ and $\mathbf{r}_a$, separately over the two-stage optimization process. 
First, $\mathbf{R}^I_L$  is obtained by minimizing the residual $\mathbf{r}_w$, and then $\mathbf{p}^I_L$ is determined by minimizing the residual $\mathbf{r}_a$ in \cite{zhu2022robust}. However, this approach overlooks the interaction and correlation between these constraints. Specifically, the residual $\mathbf{r}_a$  depends on not only $\mathbf{p}^I_L$  but also  $\mathbf{R}^I_L$. simultaneously. In our \textit{Gril-Calib}, we address this ineffectiveness by reformulating the calibration as a Single Optimization (SO). Our optimization problem could be depicted as follows,
\begin{equation}
\begin{gathered}
\mathbf{x}^*_{cal}=\underset{\mathbf{x}}{\arg \min } \sum \{ \rho_{w} \lVert \mathbf{r}_w (\mathbf{R}^I_L, \mathbf{b}_g, \mathbf{t}_f ) \rVert^2_{\Sigma_{\mathbf{r}_w}} + 
\\
\rho_{a} \lVert \mathbf{r}_a (\mathbf{R}^I_L,\mathbf{p}^I_L, \mathbf{b}_a) \rVert^2_{\Sigma_{\mathbf{r}_a}} + \rho_{g}\lVert \mathbf{r}_g (\mathbf{R}^I_L, \mathbf{p}^I_L) \rVert^2_{\Sigma_{\mathbf{r}_g}} \}
\end{gathered}
\end{equation}
and minimize the two residuals $\mathbf{r}_w$  and  $\mathbf{r}_a$ along with our proposed residual $\mathbf{r}_g$ simultaneously, where $\Sigma_{\mathbf{r}_w}, \Sigma_{\mathbf{r}_a}, \Sigma_{\mathbf{r}_g}$ denotes the Mahalanobis distance for each covariance matrix. $\rho_{x}$, where $x=w, a, g$ refers to the scaled Cauchy kernel function assigned to each residual. 
To determine the scale parameters for the Cauchy kernel function, we utilize a grid search in a predefined parameter space. We found the optimal parameter combination of  $\rho_w : \rho_a : \rho_g  = 10:1:5$ that was used consistently across all experiments.

\begin{sloppypar}
\begin{table*}[t]
\begin{threeparttable}
\centering
\caption{RMSE result compared to existing methods on M2DGR dataset}
\scriptsize 
\setlength{\tabcolsep}{2pt} 
\begin{tabular}{@{}ccccccccccc@{}}
\toprule
\multirow{2}{*}{M2DGR \cite{yin2021m2dgr}} & \multicolumn{2}{c}{\textit{\begin{tabular}[c]{@{}c@{}}Gate 01\\ {[}0,140{]}\end{tabular}}} & \multicolumn{2}{c}{\textit{\begin{tabular}[c]{@{}c@{}}Rotation 02\\ {[}0, 120{]}\end{tabular}}} & \multicolumn{2}{c}{\textit{\begin{tabular}[c]{@{}c@{}}Hall 04\\ {[}0, 160{]}\end{tabular}}} & \multicolumn{2}{c}{\textit{\begin{tabular}[c]{@{}c@{}}Lift 04\\ {[}0, 80{]}\end{tabular}}} & \multicolumn{2}{c}{\textit{\begin{tabular}[c]{@{}c@{}}Street 08\\ {[}20, 130{]}\end{tabular}}} \\ \cmidrule(l){2-11} 
 & Rotation (\textdegree) & Translation (m) & Rotation (\textdegree) & Translation (m) & Rotation (\textdegree) & Translation (m) & Rotation (\textdegree) & Translation (m) & Rotation (\textdegree) & Translation (m) \\ \midrule
ILC \cite{mishra2021target} & 29.549 & 1.210 & 11.581 & 0.839 & 48.628 & 7.517 & 9.998 & 0.390 & 7.205 & 0.847 \\
FAST-LIO2 \cite{xu2022fast} & 3.388 & 0.392 & 3.686 & 0.394 & 3.325 & 0.387 & 3.667 & 0.401 & 3.335 & 0.385 \\
OA-LICalib \cite{lv2022observability} & 0.642 & 0.237 & 0.922 & 0.208 & 0.851 & 1.742 & 0.902 & \textbf{0.021} & 0.770 & 0.241 \\
LI-Init \cite{zhu2022robust} & 1.120 & 0.383 & 1.305 & 0.311 & 2.250 & 0.287 & 3.083 & 0.295 & 0.784 & 0.223 \\
GRIL-Calib (Proposed) & \textbf{0.501} & \textbf{0.039} & \textbf{0.878} & \textbf{0.021} & \textbf{0.559} & \textbf{0.034} & \textbf{0.873} & 0.085 & \textbf{0.639} & \textbf{0.125} \\ \bottomrule
\end{tabular}
\label{table1}
\begin{tablenotes}
\item {The best-performing algorithms are shown in \textbf{bold}.}
\item {The following numbers under the sequence represent segments of data.}
\end{tablenotes}
\end{threeparttable}
\end{table*}
\end{sloppypar}

For data accumulation, we start collecting once the robot is ready after initializing the Madgwick filter. The end time for data collection is determined similarly to \cite{zhu2022robust}, ensuring the robot moves enough for accurate calibration. However, unlike \cite{zhu2022robust}, we only consider the largest singular value of the Jacobian matrix $\mathbf{J_R}$ of $\mathbf{R}^I_L$ since the ground robot is attached to the ground and moves only in planar motion.

\section{Experiments}
\label{experiments}

\subsection{Experiments Setup}

Our implementation is based on C++, using ROS on Ubuntu 18.04. We perform our experiments using a 4-core Intel® Core™ i5-6600 CPU and Titan XP GPU. We use Ceres-solver \footnote{\url{http://ceres-solver.org/}} to compute the non-linear optimization. We evaluated the algorithms using various datasets with flat ground.
Following the evaluation scheme used in \cite{zhu2022robust}, we only evaluate spatial extrinsic calibration parameters in terms of RMSE (Root Mean Squared Error), skipping the evaluation on IMU-LiDAR time offset $t^I_L$.  To show the effectiveness of our \textit{Gril-Calib}, we compare our method with other recent IMU-LiDAR calibration methods which are named ILC \cite{mishra2021target}, FAST-LIO2 \cite{xu2022fast}, OA-LICalib \cite{lv2022observability}, LI-Init \cite{zhu2022robust}. All algorithms are reproduced in the same environment.

\subsection{Extrinsic Calibration Performance on the M2DGR}
\label{m2dgr-seq}

\begin{figure}[t]
 \centering
 \begin{subfigure}[b]{0.45\textwidth}
     \centering
\includegraphics[width=0.65\textwidth]{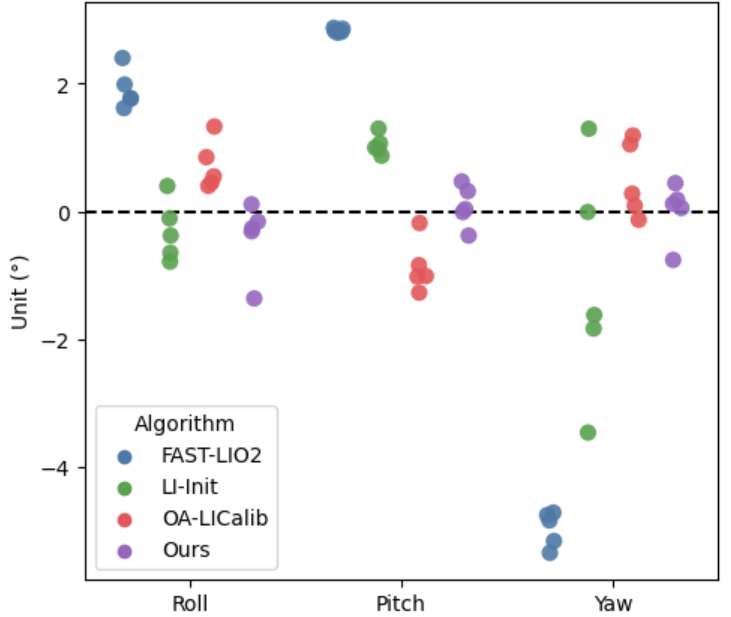}
      \caption{IMU-LiDAR extrinsic rotation parameters results}
         \label{fig4a}
     \end{subfigure}
     \vfill
     \vspace{0.2cm}
 \begin{subfigure}[b]{0.45\textwidth}
         \centering
\includegraphics[width=0.66\textwidth]{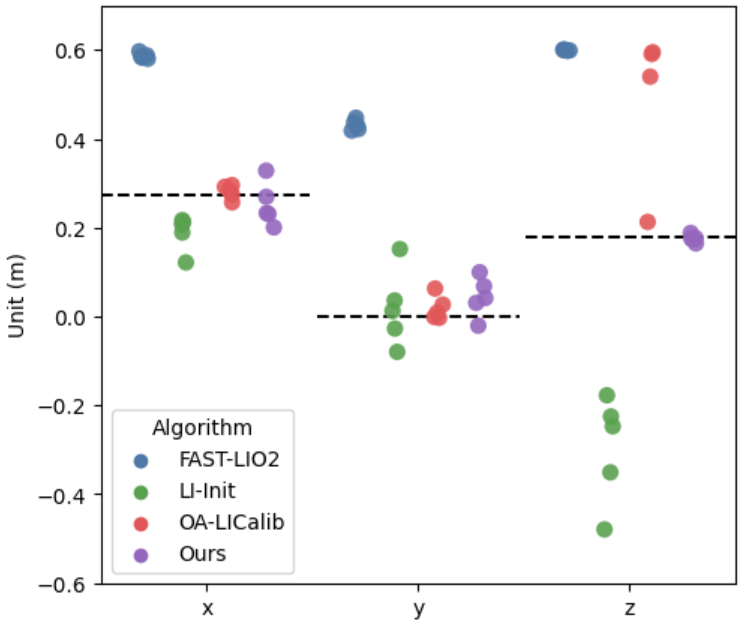}
  \caption{IMU-LiDAR extrinsic translation parameter results}
 \label{fig4b}
 \end{subfigure}
\caption{Visualization for each of the extrinsic parameters. The dotted line represents the ground truth. Closer to the dotted line represents the best estimation.}
\label{fig4}
\end{figure}

M2DGR uses Velodyne VLP-32C as the LiDAR and Handsfree A9 as the IMU. For the initial parameter settings, we decide to set values that are far enough away from the ground truth.
The initial rotation parameters are set to $\mathbf{R}^I_L = (-5.0,-5.0,5.0)$ using Euler angle conversion, with units in degrees. The initial translation parameters deviated by 0.402 meters from the ground truth based on RMSE, represented as $\mathbf{p}^I_L = (0.6,0.45,0.6)$. We initialized all remaining state vectors  $\mathbf{t}_f, \mathbf{b}_g, \mathbf{b}_a$ to zeros. 
Data was collected from both indoor (\textit{Hall 04}  and \textit{Lift 04}) and outdoor (\textit{Gate 01}, \textit{Rotation 02}, and \textit{Street 08}) environments. The calibration result on the M2DGR is summarized in Table \ref{table1}.  The table shows that our \textit{Gril-Calib} significantly outperforms the other methods. The extrinsic parameters of each calibration method are visualized in Fig. \ref{fig4}. Fig. \ref{fig4a} and Fig, \ref{fig4b} are the results of the rotation and translation matrices, respectively. In both figures, the dotted lines denote the ground truth of the extrinsic parameters. Thus, the closer to the dotted line, the better the calibration method is. ILC \cite{mishra2021target} is excluded from the figures because its error is significantly larger than that of the other methods. Additionally, we report the computation time of the algorithm to evaluate its efficiency in the supplementary material.

\begin{sloppypar}

\end{sloppypar}

\begin{table}[t]
\centering
\caption{RMSE result compared to existing methods on S3E dataset}
\scriptsize 
\setlength{\tabcolsep}{3pt} 

\begin{tabular}{@{}ccccc@{}}
\toprule
\multirow{2}{*}{S3E \cite{feng2022s3e}} & \multicolumn{2}{c}{\textit{\begin{tabular}[c]{@{}c@{}}Lab 01 $(\beta)$\\ {[}65, 210{]}\end{tabular}}} & \multicolumn{2}{c}{\textit{\begin{tabular}[c]{@{}c@{}}Lab 02 $(\beta)$\\ {[}220, 340{]}\end{tabular}}} \\ \cmidrule(l){2-5} 
 &  Rotation (\textdegree) & Translation (m) & Rotation (\textdegree) & Translation (m) \\ \midrule
ILC \cite{mishra2021target} & 6.430 & 0.288 & 9.817 & 0.813 \\
FAST-LIO2 \cite{xu2022fast} & 3.489 & 0.303 & 3.351 & 0.300 \\
OA-LICalib \cite{lv2022observability} & \textbf{0.339} & 0.194 & 18.512 & 0.321 \\
LI-Init \cite{zhu2022robust} & 1.238 & 0.116 & 1.657 & 0.066 \\
GRIL-Calib (Proposed) & 0.935 & \textbf{0.036} & \textbf{0.847} & \textbf{0.027} \\ \bottomrule
\end{tabular}
\label{table2}
\end{table}

\begin{table}[t]
\begin{threeparttable}
\centering
\caption{RMSE result compared to existing methods on HILTI dataset}
\scriptsize 
\setlength{\tabcolsep}{3pt} 
\begin{tabular}{@{}ccccc@{}}
\toprule
\multirow{2}{*}{HILTI \cite{helmberger2022hilti}} & \multicolumn{2}{c}{\textit{\begin{tabular}[c]{@{}c@{}}Basement 3\\ {[}85, 200{]}\end{tabular}}} & \multicolumn{2}{c}{\textit{\begin{tabular}[c]{@{}c@{}}Basement 4\\ {[}75, 225{]}\end{tabular}}} \\ \cmidrule(l){2-5} 
 &  Rotation (\textdegree) & Translation (m) &  Rotation (\textdegree) & Translation (m) \\ \midrule
ILC \cite{mishra2021target} & 12.496 & 1.622 & 44.294 & 4.052 \\
FAST-LIO2 \cite{xu2022fast} & 2.533 & 0.206 & 1.892 & 0.188 \\
OA-LICalib \cite{lv2022observability} & 3.979 & 0.448 & 2.368 & 0.185 \\
LI-Init \cite{zhu2022robust} & 1.652 & 0.191 & 2.584 & 0.194 \\
GRIL-Calib (Proposed) & \textbf{0.069} & \textbf{0.018} & \textbf{0.589} & \textbf{0.047} \\ \bottomrule
\end{tabular}
\label{table3}
\end{threeparttable}
\end{table}

\subsection{Extrinsic Calibration Performance on the Others}

S3E and HILTI data are used as additional real-world datasets, and their results are summarized in Table \ref{table2} and \ref{table3}. Both of the datasets are captured indoors with flat ground. The S3E dataset uses a Velodyne VLP-16 and Xsens MTi-30 IMU, while the HILTI dataset uses an Ouster OS0-64 LiDAR and AlphaSense IMU. 
Similar with Section \ref{m2dgr-seq}, we set the initial value settings as follows. In HILTI, $\mathbf{R}^I_L = (-179.91, -1.0, -5.0)$ and $\mathbf{p}^I_L = (-0.1, -0.4, 0.5)$. In S3E, $\mathbf{R}^I_L = (0.0, 0.0, 5.0)$ and $\mathbf{p}^I_L = (-0.1, -0.2, -0.4)$ respectively. As with other experiments, we initialized all remaining state vectors to zeros. As shown in Tables \ref{table2} and \ref{table3}, our algorithm outperforms other competing methods and it shows the reliable accuracy of our \textit{Gril-Calib} across various scenarios. Due to the paper limitation, experiments on the custom dataset, tests conducted on rough terrains, and the application of extrinsic calibration to the IMU-LiDAR Fusion System, are provided in supplementary material.
\footnote{\url{https://github.com/Taeyoung96/GRIL-Calib}}

\subsection{Robustness to the Initial Guess}

\begin{figure}[t]
 \centering
 \begin{subfigure}[b]{0.4\textwidth}
     \centering
\includegraphics[width=0.8\textwidth]{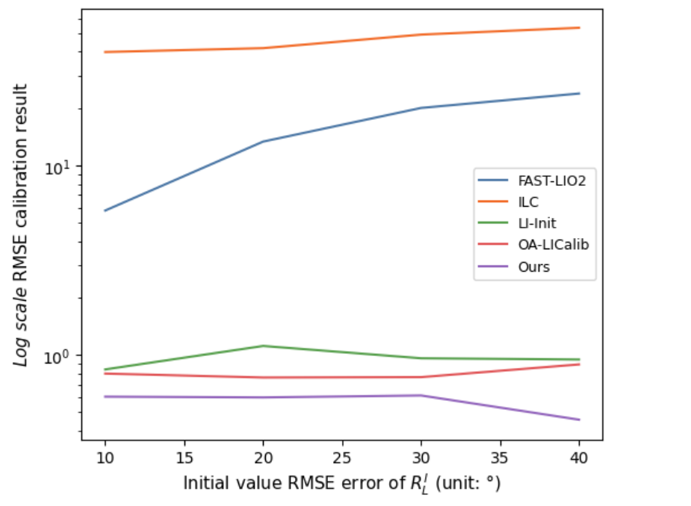}
      \caption{RMSE result of IMU-LiDAR extrinsic rotation parameters $\mathbf{R}^I_L$}
         \label{fig6a}
     \end{subfigure}
     \vfill
     \vspace{0.2cm}
 \begin{subfigure}[b]{0.4\textwidth}
 \centering
\includegraphics[width=0.82\textwidth]{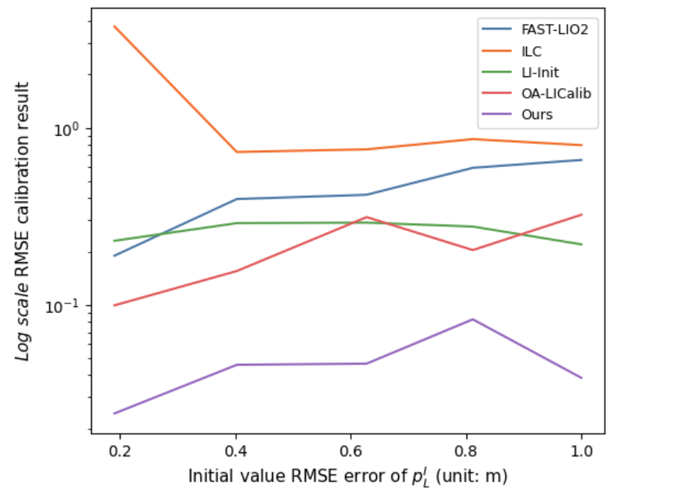}
  \caption{RMSE result of IMU-LiDAR extrinsic translation parameters $\mathbf{p}^I_L$}
 \label{fig6b}
 \end{subfigure}
\caption{Calibration performance with changes initial guess.}
\label{fig6}
\end{figure}

The extrinsic calibration usually requires initial parameter values, and it significantly affects the performance \cite{xu2023observability}. We conduct some experiments while varying initial guesses to see how robust the competing calibration methods are to the initial guesses. In our experiments, we set the initial values to be sufficiently far from the ground truth. We utilize three different sequences, \textit{Gate 01}, \textit{Rotation 02}, and \textit{Hall 04}, and average them.  We progressively set the initial values that are farther from the ground truth to visualize the results of different algorithms for changing the initial setting. The results are shown in Fig  \ref{fig6}. Fig \ref{fig6a} shows the change of the calibration error while fixing the initial $\mathbf{p}_{L}^{I}$  but changing initial $\mathbf{R}_{L}^{I}$. Fig  \ref{fig6b} is the result when $\mathbf{R}_{L}^{I}$  is fixed but $\mathbf{p}_{L}^{I}$ is varied. We observe that the accuracy of our \textit{GRIL-Calib} is less affected by the choice of initial parameters.

\begin{table}[h]
\centering
\caption{Ablation study for proposed method}
\scriptsize 
\setlength{\tabcolsep}{5pt} 
\begin{tabular}{@{}cccccc@{}}
\toprule
 & \multirow{2}{*}{GPM} & \multirow{2}{*}{SO} & \multirow{2}{*}{\begin{tabular}[c]{@{}c@{}}LO \\ w/ GP\end{tabular}} & \multicolumn{2}{c}{RMSE} \\ \cmidrule(l){5-6} 
 &  &  &  & Rotation (\textdegree) & Translation (m) \\ \midrule
{\begin{tabular}[c]{@{}c@{}}LI-Init \cite{zhu2022robust} \\ (Baseline)\end{tabular}}  &  &  &  & 1.710 & 0.300 \\ \midrule
\multirow{6}{*}{GRIL-Calib} & \Checkmark &  &  & 1.421 & 0.088 \\
 \multirow{6}{*}{(Proposed)} &   & \Checkmark  &  & 0.985 & 0.216 \\
 &  &  & \Checkmark & 1.203 & 0.213 \\
 & \Checkmark  & \Checkmark  &  & 0.695 & 0.067 \\
 & \Checkmark &  & \Checkmark & 1.109 & 0.073 \\
 &  & \Checkmark & \Checkmark & 0.914 & 0.204 \\
 & \Checkmark & \Checkmark & \Checkmark & \textbf{0.690} & \textbf{0.061} \\ \bottomrule
\end{tabular}
\begin{tablenotes}
\item { \quad $\cdot$ GPM: Add Ground Plane Motion Constraints}
\item { \quad $\cdot$ SO: Single Optimization Process}
\item { \quad $\cdot$ LO w/ GP: LiDAR Odometry with proposed Ground Plane Residual}
\end{tablenotes}
\label{table4}
\end{table}

\begin{figure}[b]
\centering
\includegraphics[width=0.7\columnwidth]{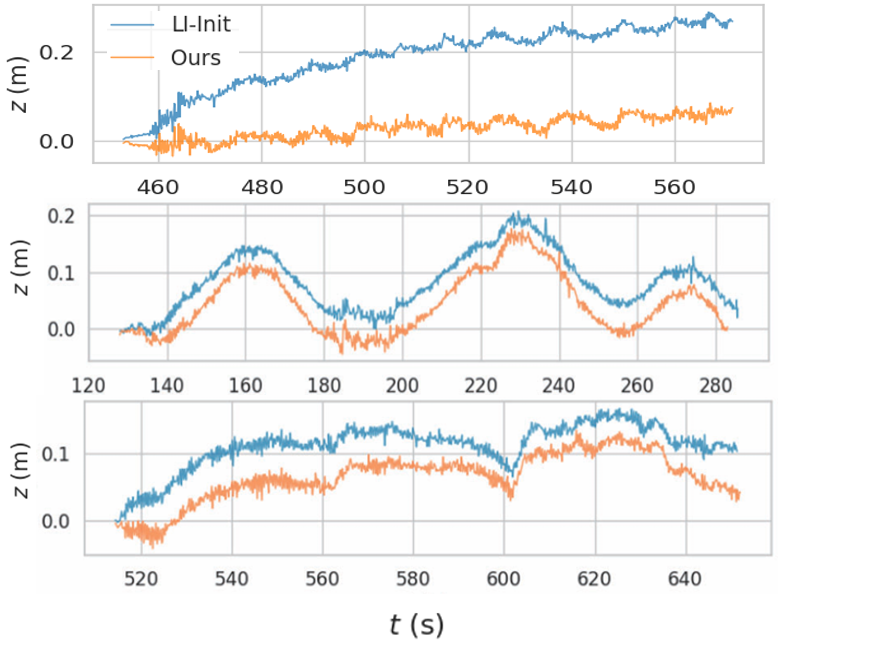}
\caption{Comparative visualization of the z-component of a trajectory using \textit{Rotation 02}, \textit{Hall 04}, \textit{Gate 01} sequence in M2DGR. (Top to bottom)}
\label{z_drift}
\end{figure}

\subsection{Ablation Study}

We ablate three key modules comparing the result with that of our baseline \cite{zhu2022robust} using whole M2DGR sequences. The results are summarized in Table \ref{table4}. 
Adding each module sequentially improved calibration performance. Specifically, the incorporation of `GPM' resulted in improvements of 16.9\% and 70\% in the rotation and translation parameters, respectively, compared to the \cite{zhu2022robust}. The GPM is particularly effective for the translation parameter in the unobservable direction, resulting in significant performance enhancements. Further improvements were observed when calibration is performed by adding Single Optimization (`SO'), with increases of 51\% in rotation and 23\% in translation parameters. This improvement underscores the high correlation between $\mathbf{R}^I_L$ and $\mathbf{p}^I_L$, and highlights that separate optimization, which overlooks this correlation, compromises accuracy. Finally, adding the ground plane residual in LiDAR odometry (`LO w/ GP') further enhances performance. The reduction in LiDAR Odometry drift with the addition of ground plane residual is shown in Fig \ref{z_drift}. The reduced drift in the odometry allows for more accurate LiDAR angular velocity and acceleration, which contributed to the improved accuracy of the calibration.

\section{CONCLUSIONS}
\label{conclusions}

In this paper, we present \textit{GRIL-Calib}: a novel targetless IMU-LiDAR extrinsic calibration framework that mainly focuses on the ground robot. Our framework aimed to address the challenge of obtaining accurate 6-DoF extrinsic parameters including unobservable direction when the robot's motion is limited to a planar motion. To achieve this, we proposed the Ground Plane Motion (GPM) constraint to utilize ground observation, which is easily accessible for the ground robots. We also reformulate the calibration as a Single Optimization (SO) problem to take advantage of the correlation of the residuals. Additionally, we proposed ground plane residual to reduce the z-drift of the LiDAR odometry, thereby improving the accuracy of the calibration. Finally, we demonstrated the effectiveness and reliability of our \textit{GRIL-Calib} by applying it to various datasets. We acknowledge that our algorithm's performance might be affected by uneven ground. Future works should focus on calibration for such terrains.

\bibliographystyle{IEEEtran}
\bibliography{bib}

\end{document}